\def\cvprPaperID{0000}
\def\confName{CVPR}
\def\confYear{2023}

\def\paperTitle{External Reasoning : Towards Multi-Large-Language-Models \\ Interchangeable Assistance with Human Feedback}

\def\authorBlock{
    Akide Liu \qquad \\
    The University of Adelaide, Australia \\
    {\tt\small akide.liu@adelaide.edu.au} \\
   	{\tt\small UI Preview: \url{https://vmv.re/chatpdf-demo}} \\
    {\tt\small Source Code: \url{https://github.com/AkideLiu/ANLP}}
}

\newif\ifreview 
\newif\ifarxiv \newcommand{\arxiv}{\arxivtrue}
\newif\ifcamera 
\newif\ifrebuttal

\arxiv 

\pdfoutput=1
\documentclass[10pt,twocolumn,letterpaper]{article}
\ifreview \usepackage[review]{cvpr} \fi
\ifarxiv \usepackage[pagenumbers]{cvpr} \fi
\ifrebuttal \usepackage[rebuttal]{cvpr} \fi
\ifcamera \usepackage{cvpr} \fi

\usepackage{graphicx}
\usepackage{amsmath}
\usepackage{amssymb}
\usepackage{booktabs}
\usepackage{algorithm}
\usepackage{algpseudocode}
\usepackage{tabularx}  
\usepackage{lscape}
\usepackage{longtable}
\usepackage{supertabular}

\usepackage{times}
\usepackage{microtype}
\usepackage{epsfig}
\usepackage[table,xcdraw]{xcolor}
\usepackage{caption}
\usepackage{float}
\usepackage{placeins}
\usepackage{color, colortbl}
\usepackage{stfloats}
\usepackage{enumitem}
\usepackage{tabularx}
\usepackage{xstring}
\usepackage{multirow}
\usepackage{xspace}
\usepackage{url}
\usepackage{subcaption}
\usepackage{xcolor}
\usepackage[hang,flushmargin]{footmisc}

\ifcamera \usepackage[accsupp]{axessibility} \fi

% ...

\ifarxiv  \fi

\newcommand{\R}[1]{{%
    \textbf{%
        \ifstrequal{#1}{1}{\textcolor{red}{R#1}}{%
        \ifstrequal{#1}{2}{\textcolor{blue}{R#1}}{%
        \ifstrequal{#1}{3}{\textcolor{magenta}{R#1}}{%
        \ifstrequal{#1}{4}{\textcolor{teal}{R#1}}{%
                           \textcolor{cyan}{R#1}%
        }}}}%
    }%
}}

\usepackage{xr-hyper}

\makeatletter
\newcommand*{\addFileDependency}[1]{
  \typeout{(#1)}
  \@addtofilelist{#1}
  \IfFileExists{#1}{}{\typeout{No file #1.}}
}

\makeatother

\usepackage[pagebackref,breaklinks,colorlinks]{hyperref}
\usepackage[capitalize]{cleveref}
\crefname{section}{Sec.}{Secs.}
\crefname{table}{Table}{Tables}
\crefname{figure}{Fig.}{Figs.}

\begin{document}
\title{\paperTitle}
\author{\authorBlock}

\twocolumn[{
\renewcommand\twocolumn[1][]{#1}%
\maketitle
\begin{center}
 \centering
 \captionsetup{type=figure}
 \includegraphics[width=0.9\textwidth]{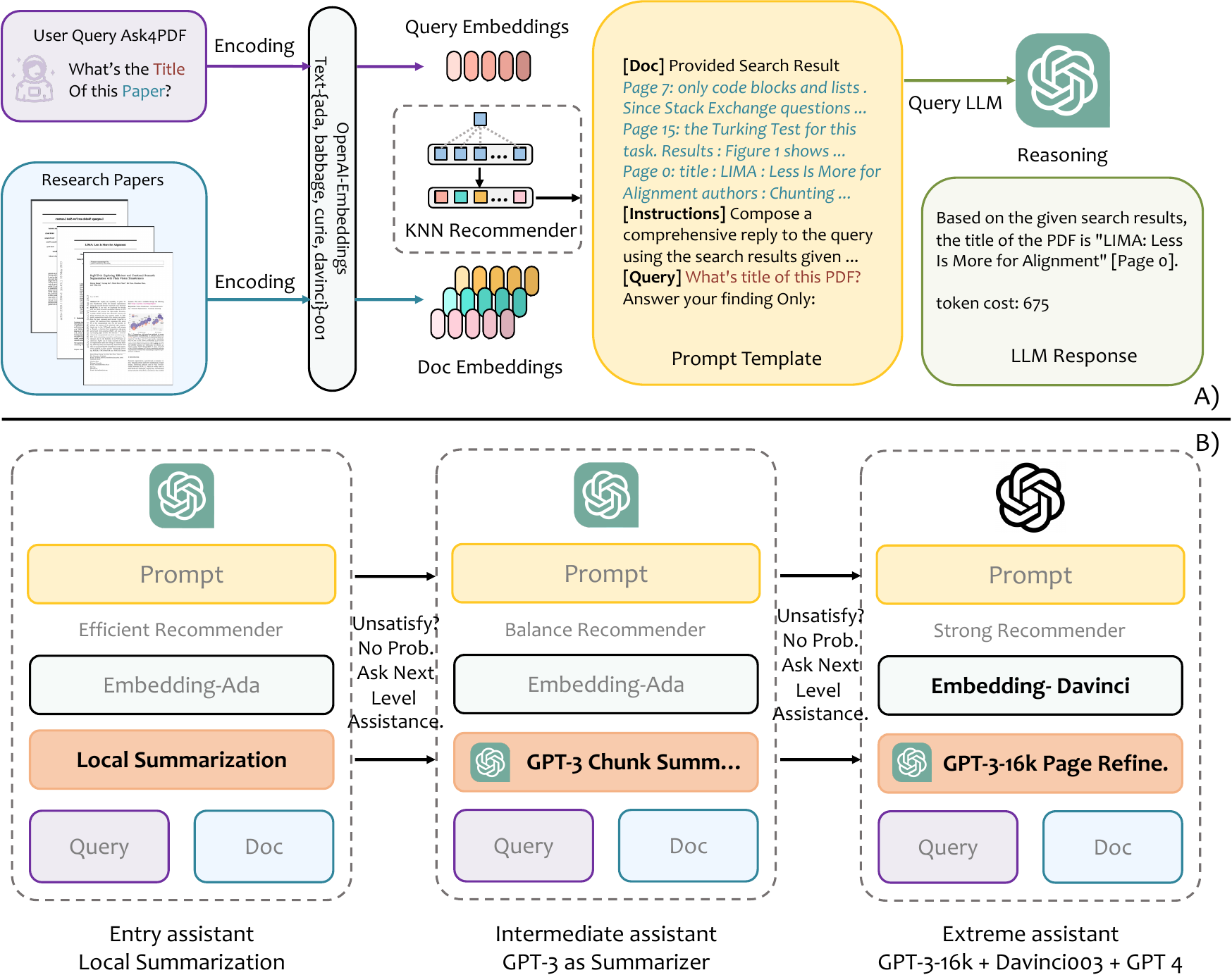}

	\caption{
	Towards Multi-Large-Language-Models Interchangeable Assistance with Human Feedback System Overall Framework.
In \textit{Section A)}, we elucidate the operational workflow of the Chat Research Paper System. The process commences when a user submits a query concerning a specific research paper. The system initially retrieves the PDF of the paper from an online repository or, alternatively, from local storage if available. The document is then segmented into multiple portions which undergo an embedding process using OpenAI Embeddings, generating distinct query and document embeddings. Following this, a k-nearest neighbors algorithm is employed to match these embeddings, facilitating the extraction of the most relevant segments of the document in relation to the user’s query. These pertinent segments, along with instructions and the user's query, are then submitted to the Large Language Model (LLM) for processing.
In \textit{Section B)}, we expound on the policy-oriented aspect of the system, which dynamically adjusts its approach based on human feedback, particularly when ambiguous queries or unsatisfactory responses from the LLM are encountered. Users have the option to escalate the level of assistance rendered by the system. At the entry level, the system employs an efficient matching algorithm and integrates a local summarization layer for document embeddings. Ascending to the intermediate level, GPT-3 is engaged to perform chunk-level filtering and summarization. At the most advanced stage, the system relaxes token window limitations and utilizes GPT-3-16k to conduct multi-page retrieval and refinement. During this phase, the embeddings are generated using the cutting-edge Davini003 model, and a robust recommendation system is employed to craft a comprehensive prompt that queries GPT-4 for sophisticated reasoning. Additionally, all intermediate outputs are retained in a memory buffer for potential utility in addressing subsequent queries.
	}
 \label{fig:overall}
\end{center}
}]

\maketitle

\begin{abstract}

Memory is identified as a crucial human faculty that allows for the retention of visual and linguistic information within the hippocampus and neurons in the brain, which can subsequently be retrieved to address real-world challenges that arise through a lifetime of learning. The resolution of complex AI tasks through the application of acquired knowledge represents a stride toward the realization of artificial general intelligence. However, despite the prevalence of Large Language Models (LLMs) like GPT-3.5 and GPT-4 \cite{brown2020language, leiter2023chatgpt, zaitsu2023distinguishing, OpenAI2023GPT4TR} , which have displayed remarkable capabilities in language comprehension, generation, interaction, and reasoning, they are inhibited by constraints on context length that preclude the processing of extensive, continually evolving knowledge bases. This paper proposes that LLMs could be augmented through the selective integration of knowledge from external repositories, and in doing so, introduces a novel methodology for External Reasoning, exemplified by ChatPDF. Central to this approach is the establishment of a tiered policy for \textbf{External Reasoning based on Multiple LLM Interchange Assistance} in \cref{fig:overall}, where the level of support rendered is modulated across entry, intermediate, and advanced tiers based on the complexity of the query, with adjustments made in response to human feedback. A comprehensive evaluation of this methodology is conducted using multiple LLMs and the results indicate state-of-the-art performance in \cref{comparison} , surpassing existing solutions including ChatPDF.com. Moreover, the paper emphasizes that this approach is more efficient compared to the direct processing of full text by LLMs. The source code is public available at : \url{https://github.com/AkideLiu/ANLP} .

\end{abstract}
\section{Introduction}
\label{sec:intro}


\begin{figure}[hbt]
\resizebox{\columnwidth}{!}{%
  \includegraphics{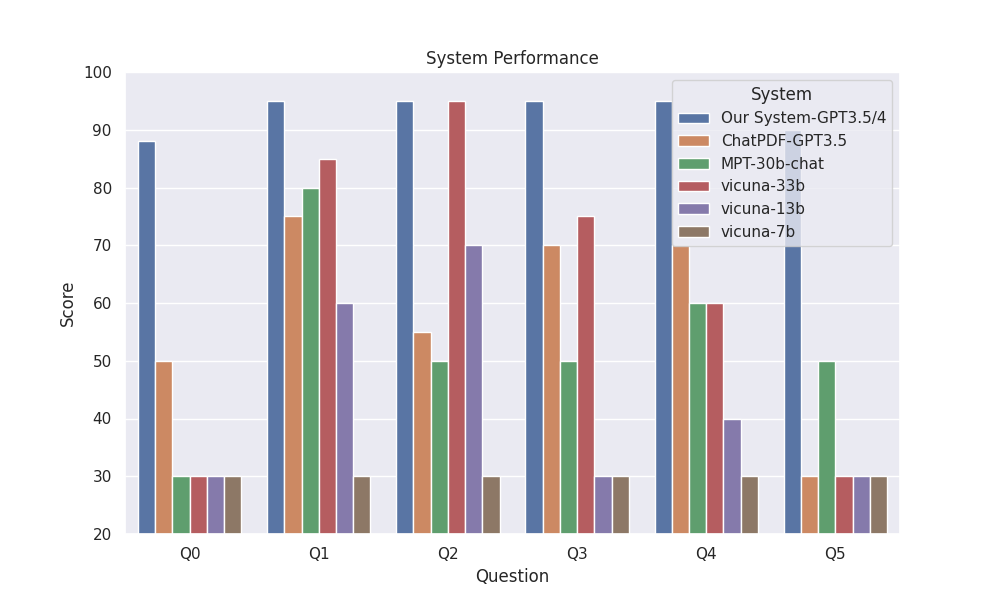}}
  \caption{Performance Comparison of various systems, including ChatPDF.com. The evaluation process is carried out using a model called Claude+, which employs a super-large context window length of 100K tokens. During the evaluation, the entire PDF document is utilized as input, and the performance is examined using different candidate models. To further enhance the evaluation, our proprietary solutions are replaced with open-source large language models such as MPT-30B-chat, Vicuna-33B, Vicuna-13B, and Vicuna-7B. These models are loaded locally using Full precision mode with the assistance of 4 A100 GPUs. The Evaluation shows our system achieves SOTA.}
  \label{comparison}
\end{figure}

Large Language Models (LLMs) \cite{brown2020language, Ouyang2022TrainingLM, Chowdhery2022PaLMSL, Zhang2022OPTOP, Zeng2022GLM130BAO, Touvron2023LLaMAOA} such as ChatGPT have garnered widespread interest from both scholarly and commercial arenas, attributable to their extraordinary efficacy across a gamut of Natural Language Processing (NLP) tasks. These LLMs are architected upon an edifice of extensive pre-training on colossal text corpora, complemented by the augmentation of Reinforcement Learning from Human Feedback (RLHF) \cite{Ouyang2022TrainingLM} . This dual-pronged approach imbues LLMs with an unparalleled capacity for language comprehension, generation, interaction, and reasoning.
The formidable capabilities of LLMs have catalyzed a proliferation of nascent research domains that are poised to explore and harness the immense potential intrinsic to these models. Notably, these burgeoning areas of exploration include in-context learning \cite{brown2020language, Xie2021AnEO, Min2022RethinkingTR}, instruction-based learning  \cite{Wei2021FinetunedLM, Wang2022SuperNaturalInstructionsGV, Iyer2022OPTIMLSL, Chung2022ScalingIL} , and chain-of-thought prompting \cite{Wei2022ChainOT, Kojima2022LargeLM, Gao2022PALPL, Wang2022SelfConsistencyIC} , amongst others. These emergent research directions exemplify the seemingly boundless prospects that LLMs afford in the quest to architect sophisticated artificial intelligence systems.
This work, in particular, ventures into External Reasoning as an avenue to exploit the strengths of LLMs through a system that not only incorporates the interchangeability of various LLMs but also judiciously integrates human feedback, fostering a symbiosis between human intelligence and computational language models in the pursuit of advanced reasoning capabilities.

Despite the remarkable achievements attributed to Large Language Models (LLMs), these technologies harbor intrinsic limitations that pose substantial challenges in the development of advanced AI systems. We analyze these challenges from three distinct vantage points:
1) Static Knowledge Post-Training: LLMs acquire their knowledge reservoir through unsupervised pre-training stages. However, subsequent to this phase, the knowledge base of LLMs becomes immutable. Incorporating or updating the knowledge is computationally prohibitive due to the extensive resources required, which stifles the models' adaptability and responsiveness to evolving information landscapes.
2) Inadequacies in Zero-Shot Learning: LLMs may exhibit suboptimal performance in zero-shot learning settings, even when they have encountered relevant information during the pre-training stage. This is attributed to the disproportionate weighting of data during training, which can hinder the models’ ability to effectively recall and apply pertinent knowledge when required.
3) Context Window Limitations in Few-Shot Settings: In scenarios that utilize few-shot learning, LLMs are constrained by the finite context window length, which restricts the amount of information that can be processed in a single interaction. Furthermore, the efficient retrieval of highly relevant text sources from a local context poses a formidable challenge, as the models must discern and extract the most germane information from a constrained dataset.
Addressing these challenges necessitates innovative approaches that enhance the flexibility and adaptability of LLMs, optimize data weighting strategies for improved zero-shot performance, and develop efficient retrieval mechanisms that can operate within the context window constraints. This would pave the way for more robust and versatile AI systems. 

In this paper, we underscore the challenge of incorporating ongoing knowledge and external data, sourced from external retrieval systems, into the processing pipeline of chat-based PDF document interaction systems, such as ChatPDF. The crux of this challenge lies in the development of an efficacious external filter system capable of extracting the most pertinent information to address the queries at hand. 
As an initial step, we establish a baseline methodology that entails the matching of query and document embeddings to identify the most relevant sections within the document. Subsequently, a prompt template is formulated, incorporating the document, instructions, and query, to solicit reasoning from GPT-3.5, as depicted in Section A of Figure (refer to \cref{fig:overall}). However, a comprehensive evaluation reveals that the baseline system is adept at addressing Type 1 level questions, which involve explicit keyword matching, but is inept at handling the more nuanced Type 2 level questions that demand the extraction of implicit information such as main insights.

To surmount this limitation, we introduce a novel, policy-oriented algorithm titled "\textbf{Multi-Large-Language-Models Interchangeable Assistance with Human Feedback}". This innovative system fosters interaction between localized specialized models and LLMs, executing a preliminary stage of processing that encompasses summarization, refinement, and document similarity retrieval shows in \cref{fig:overall} secion B. The system is stratified into three tiers: entry, intermediate, and extreme. These tiers offer escalating levels of support to cater to the complexity of user queries. Specifically, in instances where users encounter challenging questions, the system offers the flexibility to elevate the level of assistance, thereby adapting to the users’ requirements.
It is imperative to highlight that our approach endeavors to strike a judicious balance between response quality and resource utilization. At the entry-level, the system is designed to efficiently address less complex tasks using a minimal token count. Conversely, at the extreme level, the system adopts a more resource-intensive approach, transcending cost constraints in a bid to deliver the highest quality responses that are commensurate with user expectations. This tiered approach ensures agility and adaptability while optimizing resource allocation. 

Additionally, we propose a solution to a compelling task (Type 3) - identifying the key references within a paper. Since key references often denote seminal or influential works in a given field, ascertaining these references empowers users with insights into the genesis and evolution of the field. In addressing this task, we present a memory-enhanced approach.
Elaborating further, when a user seeks to uncover key references, it is presumed that the user has engaged in multiple interactions with the system. Throughout this course of interactions, our system caches the summarized results derived from the user's queries and the responses generated by the LLM. Upon receiving a query for key references, the system retrieves the paper's summary from the cache, amalgamates it with the raw abstract section of the paper, and submits this combined text to the LLM. The LLM is then tasked with classifying and identifying the references based on their titles.
We posit that in a well-constructed research paper, the author elucidates several core components, such as challenges within the field, the proposed solutions, performance evaluations, and the impact on the field. By utilizing these components as reference points, the LLM is better equipped to pinpoint key references that are indicative of groundbreaking or consequential works in the respective field. This memory-enhanced approach not only streamlines the process but also fosters a deeper understanding of the research landscape.

We assess the efficacy of our system through a comparative analysis with results generated by a fully-equipped expert system. For this purpose, we employ an examiner, an advanced super-large window LLM named Claude+ (capable of handling 100K token lengths). We input the entire PDF text into the examiner to generate responses, which we consider as the ground truth. Our rationale is that with comprehensive in-context information, the results derived from the examiner should exhibit greater stability and accuracy compared to those obtained from our system, which operates under constrained conditions involving few-shot scenarios.
It is crucial to emphasize that our evaluation is not centered on assessing the reasoning capabilities of LLMs. Rather, the focus is directed towards understanding the performance of the retrieval system and analyzing how LLMs respond when provided with limited or potentially unstable references.
In addition to evaluating our system, we also scrutinize responses from several widely-used open-source LLMs under the same challenging conditions. This enables us to gauge the comparative efficacy and robustness of different LLMs in scenarios where retrieval systems must operate with constraints.

\section{Method}
\label{sec:method}

\begin{algorithm}
\caption{Structure of a Scientific Paper}
\begin{algorithmic}[1]
\State \textbf{Structure} Paper
\State \Comment{Title of the paper}
\State $\text{Paper.title} \gets \text{string}$
\State \Comment{Abstract of the paper}
\State $\text{Paper.abstract} \gets \text{string}$
\State \Comment{Sections of the paper, each with heading and text}
\For{section in Paper.sections}
    \State $\text{section.heading} \gets \text{string}$
    \State $\text{section.text} \gets \text{string}$
\EndFor
\State \Comment{References cited in the paper}
\For{reference in Paper.references}
    \State $\text{reference.title} \gets \text{string}$
    \State $\text{reference.year} \gets \text{string}$
    \State $\text{reference.journal} \gets \text{string}$
    \State $\text{reference.author} \gets \text{string}$
\EndFor
\State \Comment{Figures included in the paper}
\For{figure in Paper.figures}
    \State $\text{figure.figure\_label} \gets \text{string}$
    \State $\text{figure.figure\_type} \gets \text{string}$
    \State $\text{figure.figure\_id} \gets \text{string}$
    \State $\text{figure.figure\_caption} \gets \text{string}$
    \State $\text{figure.figure\_data} \gets \text{string}$
\EndFor
\State \Comment{Digital Object Identifier of the paper}
\State $\text{Paper.doi} \gets \text{string}$
\end{algorithmic}
\label{alg:ds}
\end{algorithm}

\subsection{PDF parsing and Preprocessing}

In this section, we delve into the utilization of a robust PDF parser, specifically tailored for academic research, named `scipdf\_parser`. This parser is adept at dissecting scholarly articles and categorizing the extracted content into distinct segments, namely: 'title', 'authors', 'abstract', 'sections', and 'references' shows in \cref{alg:ds}.
Through the course of our experiments, an intriguing observation emerged regarding the inclusion of the 'references' segment. We discerned that omitting the 'references' section from the parsed content led to a discernible enhancement in matching accuracy for the majority of queries. The rationale behind this improvement is that the 'references' section is replete with titles from cited papers, which often bear a resemblance to the terminology utilized in the main article. However, these citations tend not to contribute substantially to the reasoning process, and, as such, their inclusion can inadvertently result in less meaningful matches.
Following the parsing process, the next step involves segmenting the content of the PDF into multiple chunks. This segmentation is instrumental for facilitating more granular processing of the text. By default, each chunk encompasses 150 tokens, which has been determined to be a judicious size for ensuring both computational efficiency and the preservation of contextual information necessary for subsequent processing stages.

\begin{table}[htbp]
  \centering
  \caption{Comparison of Embedding models on Linear probe classification over 7 datasets.}
  \label{tab:models}
  \begin{tabular}{lc}
    \toprule
    \textbf{Model} & \textbf{Accuracy (\%)} \\
    \midrule
    Prevs SOTA & 90.20 \\
    text-embedding-ada & 89.30 \\
    text-embedding-babbage & 91.10 \\
    text-embedding-curie & 91.50 \\
    text-embedding-davinci & 92.20 \\
    \bottomrule
  \end{tabular}
  \label{tab:embedding_acc}
\end{table}

\subsection{Context embeddings}

In our research, a critical component of our methodology is the utilization of OpenAI's embedding models to generate embeddings for queries and documents. The objective of this process is to enable efficient and accurate retrieval of sections that are highly pertinent to the queries at hand. 
Among the OpenAI text embedding models we employ are: text-embedding-ada, text-embedding-babbage, text-embedding-curie, and text-embedding-davinci. These models exhibit a hierarchical nature in terms of their capabilities with text-embedding-ada at the basic end of the spectrum, and text-embedding-davinci being the most sophisticated among them. As we progress along this hierarchy, the quality of the embeddings improves significantly, which can result in more accurate section retrieval show in \cref{tab:embedding_acc}
However, it is imperative to recognize that the enhanced capabilities of the more sophisticated models come at the expense of increased computational costs. Consequently, there is an inherent trade-off between the quality of embeddings and the computational overhead involved.
In light of this trade-off, judicious selection of embedding models is essential. In our system, we adopt text-embedding-ada as the default model for generating embeddings. This choice is rooted in its ability to offer a balance between performance and computational efficiency for a broad range of queries. Nevertheless, when the system is engaged in the 'extreme level assistance' mode - a setting that is triggered when faced with particularly challenging queries we opt to maximize all system components to their highest capacity. In this mode, we employ text-embedding-davinci owing to its superior capability to create high-quality embeddings. It is in this context that the system is willing to incur the additional computational costs to ensure optimal performance in addressing the complex nature of the queries under consideration.

\subsection{Context Retrieval}

We introduce two distinct methodologies for context retrieval, namely cosine similarity matching and k-nearest neighbors (KNN) based matching.
Cosine similarity matching operates by computing the cosine of the angle between two vectors, representing the document embeddings and the query embeddings. This measure quantifies the similarity between the embeddings, where a value of 1 denotes perfect alignment and 0 indicates orthogonality.
Let $d$ represent the document embedding vector and $q$ represent the query embedding vector. The cosine similarity $S_{cos}$ is calculated as follows:

\begin{equation}
S_{cos}(d, q) = \frac{d \cdot q}{\|d\| \|q\|}
\end{equation}

Where $\cdot$ denotes the dot product, and $\|\cdot\|$ represents the norm of a vector. The system ranks the chunks according to the cosine similarity scores and retrieves the chunks with the highest scores as the most relevant context.

KNN-based matching involves locating the k-nearest neighbors of the query embedding in the space of document embeddings. This is based on the assumption that similar content will have nearby embeddings.
Let $D = \{d_1, d_2, ..., d_n\}$ be the set of document embedding vectors and $q$ be the query embedding vector. We define a distance metric, commonly the Euclidean distance, between the query embedding and each document embedding. For any two vectors $a$ and $b$, the Euclidean distance $D_{E}$ is defined as:
\begin{equation}
D_{E}(a, b) = \sqrt{\sum_{i=1}^{n} (a_i - b_i)^2}
\end{equation}
The system retrieves the k document embeddings that have the smallest Euclidean distances to the query embedding. These are considered as the k-nearest neighbors and are regarded as the most relevant context for the given query.
In practice, selecting the appropriate value for k is crucial as it can impact the quality of the retrieved context. Additionally, while cosine similarity is often used for high-dimensional data, KNN can be more suitable for data where the intrinsic geometry is meaningful. The choice between these methods should consider the characteristics of the data and the application requirements.

\subsection{Prompt Engineering}

Prompt engineering is a critical area of research that directly influences the performance of Large Language Models (LLMs) and Question-Answering (QA) systems. The construction of effective prompt templates is essential for eliciting accurate and relevant responses from LLMs. We propose the following structured template for posing questions to the LLM:

\begin{enumerate}
  \item \textbf{Document Context}: We define the document context with the notation [docs], followed by a supporting prompt that presents the relevant document chunks to the LLM. The template for this part is as follows:
  
  \begin{quote}
  I will provide the document chunks as follows: [docs]
  \end{quote}
  \item \textbf{Instructional Context}: We denote the instructions for response composition with the notation [Instructions]. This part of the template guides the LLM on how to structure the response, use citations, distinguish between subjects, and ensure the accuracy and relevance of the content. The template for this part is as follows:
  \begin{quote}
  Instructions: Compose a comprehensive reply to the query using the provided document chunks. Cite each reference using [Page Number] notation (each document chunk begins with this number). Ensure citations are placed at the end of each sentence. If the document chunks mention multiple subjects sharing the same name, create separate responses for each. Include only information found in the document chunks, and refrain from adding extraneous details. Ensure the accuracy of the response and avoid disseminating false content. Exclude search results that are not pertinent to the question. Respond concisely and in a step-by-step manner. 
  \end{quote}
  \item \textbf{Query Context}: Finally, we denote the actual query with the notation [query]. This part of the template specifies the question and asks the LLM for a detailed response based on its findings. The template for this part is as follows:
  \begin{quote}
  Query: [query]. Please provide detailed findings in response to the query:
  \end{quote}
\end{enumerate}
This structured approach ensures that the LLM is provided with clear context, guided instructions, and the specific query, which collectively contribute to eliciting a well-formed, accurate, and informative response. Prompt engineering in this manner can substantially enhance the efficacy and applicability of LLMs in QA systems.

\subsection{Summarization}

\textbf{Performance Enhancement.} During our experiments, we observed that raw text chunks occasionally struggled to encapsulate semantic meanings or exhibit relevance, especially in the context of Type 2 questions, which often embody implicit meanings in conjunction with user queries. In light of this, we carried out a series of experiments to devise a mechanism that could distill the essential ideas from each text chunk through summarization.
In our approach, we leveraged summarization algorithms to condense text chunks, thereby capturing the crux of the content. For the entry-level assistance, we employed a pre-trained summarization model, specifically the "facebook/bart-large-cnn" model. This model was chosen for its proficiency in generating concise summaries and its relatively low computational overhead, making it suitable for initial, low-resource processing.
For intermediate-level assistance, we further elevated the summarization process by utilizing GPT-3.5. The incorporation of GPT-3.5 allowed for not only summarization but also refinement of the text chunks. The refinement process aimed to sharpen the focus of the summary, ensuring that the essential elements align more closely with the user's query.
By summarizing and refining text chunks before they are processed by the Large Language Model, our approach enhances the relevance and accuracy of the responses to Type 2 questions, which are characterized by implicit meanings. This method simultaneously alleviates the challenge of representing semantic meanings with raw text chunks, paving the way for more effective and efficient question-answering systems.

\textbf{Efficiency Improvement.} Another salient aspect of employing summarization in the processing pipeline is the notable enhancement in efficiency and reduction in computational costs. Large language models (LLMs), by design, are often constrained by token limits, and processing extensive text chunks may incur high computational expenses and protracted processing times. Summarization serves as a linchpin to circumvent these constraints.
By distilling text chunks into condensed summaries, the volume of data fed into the LLM is considerably reduced. This contraction in data not only ensures that the input remains within the token limits of the LLM but also decreases the computation required for processing. Consequently, this leads to faster response times, making the system more agile and efficient.
Furthermore, the reduction in computational requirements translates to diminished resource utilization, which in turn leads to lower operating costs. This is especially vital in scenarios where scalability and cost-effectiveness are essential attributes.
In addition, by focusing on the essence of the content through summarization, the LLM can allocate more computational resources to in-depth analysis and generating higher quality responses. This results in not just a cost-effective and efficient system, but one that is also more effective in addressing complex queries.
In conclusion, the integration of summarization into the processing pipeline is a multi-faceted strategy that boosts efficiency, reduces computational expenses, and fosters an environment conducive to generating higher quality responses in LLM-based question-answering systems.

\subsection{Policy Oriented Muti-LLMs assistance}

As delineated in Section B of Figure \ref{fig:overall}, a distinguishing innovation in our approach lies in the incorporation of a policy system which empowers users with the flexibility to choose between different levels of assistance, in order to address their inquiries more effectively. This feature is particularly germane in instances where the responses generated by the system may not meet the user’s expectations or requirements.
Our system, cognizant of varying complexity in user queries, operates across three tiers of assistance, each progressively employing more sophisticated strategies to render higher quality responses. The initial two tiers primarily engage summarization techniques to distil and represent text chunks succinctly. While this approach is adept at handling queries of moderate complexity, it may falter when faced with queries demanding a deeper understanding or requiring the synthesis of information scattered across extensive text. 
For such intricate scenarios, users are accorded the option through the user interface to escalate their query to a higher level of assistance. This triggers the activation of the policy system, which navigates through the available assistance levels.
In the event that a user remains dissatisfied with the outputs of the first two tiers, our system elevates the query to the 'extreme' level of assistance. At this juncture, the system harnesses the prowess of GPT-3-16K, deploying it to perform multi-page level refinement. Significantly, at this stage, there is no imposition of context length reduction, especially if the contextual information bears high relevance to the query. This ensures a thorough examination of the content, facilitating a more nuanced and comprehensive response.
This policy-driven, multi-tiered approach is instrumental in adapting the system to the diverse range of user queries, efficiently marshalling resources, and optimizing response quality in alignment with the complexity of the inquiry at hand. The system, therefore, maintains a judicious equilibrium between efficiency and depth, tailoring its approach to meet the evolving demands of the user.

\subsection{Key Reference Matching}

The process of identifying and obtaining key references within academic papers is crucial for understanding the foundational work or significant advancements in a given field. This section presents an innovative approach, termed "Key Reference Matching," which facilitates this process through memory augmentation and advanced information retrieval techniques. Our system, predicated on the hypothesis that a user engages in multiple interactions with the system, employs a memory-enhanced technique to refine the selection of key references. It operates on the premise that users may have already engaged with the system multiple times before requesting key references. The memory component of our system plays a pivotal role in augmenting the retrieval of key references. As the user interacts with the system, the responses and document summaries generated by LLMs are cached. When the user queries for key references, the system recalls the cached summaries and combines them with the raw abstract section of the paper. For the classification of references, the system integrates the recalled summary with the abstract section, and feeds the concatenated information to an LLM. The objective is to classify and identify references that are central to the research paper. This is grounded in the assumption that well-crafted research papers typically delineate the challenges in the field, proposed solutions, performance evaluation, and the impact of the field, often citing seminal works. This memory-enhanced approach yields multiple benefits. It reduces the redundancy in processing, as the cached summaries are reused, and ensures a more comprehensive understanding by combining different segments of the paper. Furthermore, by focusing on the essential components of the paper, it prioritizes the references that are most likely to be of key significance.

\section{Ablation Study}

\begin{figure}[hbt]
\resizebox{\columnwidth}{!}{%
  \includegraphics{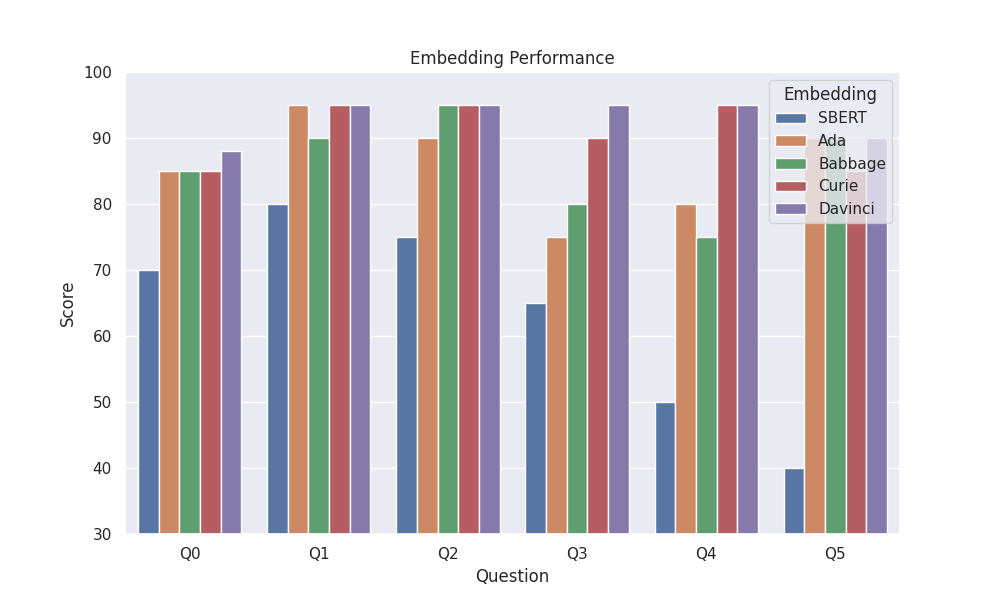}}
  \caption{Performance Comparison of various embedding systems, we evaluate the score by expert system,  Claude+ 100K.}
  \label{fig:embedded_comp}
\end{figure}

\subsection{Impact of embedding methods.} Our experiment involved a comparative analysis of Sentence-BERT (SBERT) embeddings and several OpenAI embeddings, namely, text-embedding-ada, text-embedding-babbage, text-embedding-curie, and text-embedding-davinci shows in \cref{fig:embedded_comp} . We aimed to gauge the performance of these embedding techniques across various types of questions, with particular emphasis on their capacity to accurately capture semantic relationships. The results of our experiments exhibited a clear hierarchy in performance among the assessed embedding models. It was observed that, across the spectrum of question types, the SBERT embeddings underperformed relative to their OpenAI counterparts. We attribute this disparity in performance to the smaller model size and lower dimensionality of the SBERT embeddings. Furthermore, within the cohort of OpenAI embeddings, text-embedding-curie and text-embedding-davinci stood out as superior in terms of their matching accuracy, particularly for Type 2 questions, which require a deeper understanding of semantic relationships. Conversely, text-embedding-ada and text-embedding-babbage delivered comparatively weaker performance on Type 2 questions. The choice of embedding model plays a pivotal role in determining the performance of natural language processing tasks, particularly in context retrieval. Our findings underscore the superiority of text-embedding-curie and text-embedding-davinci in tasks requiring sophisticated semantic understanding. However, it is crucial to balance the demands of the task with the computational efficiency and resource considerations.

\begin{table}
\centering
\begin{tabular}{l|cccccc} 
\toprule
\multicolumn{1}{l}{} & \multicolumn{1}{l}{Q0} & \multicolumn{1}{l}{Q1} & \multicolumn{1}{l}{Q2} & \multicolumn{1}{l}{Q3} & \multicolumn{1}{l}{Q4~} & \multicolumn{1}{l}{Q5} \\ 
\hline\hline
Cosine S=150 k=3 & 30 & 70 & 65 & 30 & 40 & 70 \\
Cosine S=150 k=5 & 85 & 75 & 80 & 50 & 30 & 80 \\
Cosine S=300 k=5 & 85 & 70 & 75 & 55 & 50 & 70 \\ 
\midrule
KNN S=150 K=3 & 85 & 80 & 75 & 60 & 70 & 80 \\
KNN S=300 K=3 & 85 & 90 & 85 & 70 & 60 & 80 \\
KNN S=300 K=5 & \textbf{85} & \textbf{95} & 70 & 90 & \textbf{95} & \textbf{90} \\
KNN S=512 K=6 & 85 & 90 & 85 & \textbf{95} & 95 & 90 \\
\bottomrule
\end{tabular}
\caption{Impact of retrieval methods. We conduct experiments with Cosine similarly match and KNN match algorithm, we evaluate the score by expert system,  Claude+ 100K. S is the size of segment chunk and K is top k answers in the algorithms.}
\label{tab:retrieve}
\end{table}

\subsection{Impact of matching algorithms.} 
In this section, we analyze the effectiveness of different matching algorithms in the retrieval of relevant document sections shows in \cref{tab:retrieve} . The primary algorithms under scrutiny are Cosine Similarity Matching and k-Nearest Neighbors (KNN) Matching. This analysis serves to inform the selection of the most effective algorithm, and the corresponding parameter configuration, for the retrieval task. As can be seen from the table, the KNN matching algorithm consistently outperformed the Cosine Similarity Matching in retrieving relevant sections. Particularly, KNN with segment size 300 and k=5 exhibited the highest performance across multiple queries. The superior performance of the KNN matching algorithm, especially with specific parameter configurations, can be attributed to its inherent ability to capture the semantic relationships more effectively than the Cosine Similarity Matching. Furthermore, by allowing flexibility in the number of nearest neighbors, KNN enables consideration of a broader context, which is particularly beneficial for complex queries.

\section{Limitations}

While we have proposed a highly flexible system, there are certain limitations to be noted. One such limitation pertains to the upgrading mechanism, which is currently reliant on human feedback. This dependence may affect the user experience. To address this, one potential solution is the integration of a sentiment analyzer that gauges the sentiment of the responses generated by the large language models (LLMs). There are instances when LLMs may fail to generate an appropriate response due to a mismatch between the local document and the user query. In such cases, LLMs might return a response like “The text provided is not related to the query”. By employing sentiment analysis, the system can discern this as a negative sentiment from the LLM and consequently escalate the query to a higher level of assistance automatically.
Another limitation arises due to the restrictive context window sizes of the LLMs, which are typically 4k or 8k tokens. This constraint can result in the system's failure to adequately address complex queries that necessitate a broader understanding of the context or the synthesis of information from multiple documents. One way to mitigate this limitation is by increasing the context window size. To this end, ALiBi, a linear-biased attention mechanism, could be integrated into the system to allow for an adjustable maximum token length at the interface level.
This proactive adaptation, through sentiment analysis and the incorporation of mechanisms like ALiBi, can potentially lead to a more fluid and effective interaction, enhancing both the system’s capabilities and the user experience. Furthermore, these adaptations emphasize the importance of the system's ability to recognize its limitations and make automatic adjustments in real-time to meet the demands of complex queries.

\section{Conclusion}

In this study, we have presented an intricate system that harnesses the capabilities of Large Language Models (LLMs) to solve complex queries, particularly in the context of retrieving and synthesizing information from scientific papers. Through a series of innovations, including sophisticated embedding methods, a novel key reference matching algorithm, and a policy system that employs varying levels of assistance, our system achieves remarkable flexibility and adaptability.
However, it is essential to recognize that with the sheer complexity and evolving nature of natural language processing, there is no one-size-fits-all solution. The limitations of the system, such as reliance on human feedback for upgrading assistance levels and the challenges posed by restrictive context window sizes of LLMs, were candidly acknowledged. We further explored potential improvements, such as the integration of sentiment analysis to autonomously escalate queries to higher levels of assistance, and employing ALiBi to permit an adjustable maximum token length.
Going forward, it is evident that as the field of natural language processing continues to evolve, systems like ours will need to continually adapt and innovate. Not only must they address the existing challenges but also stay ahead of the curve in incorporating emerging technologies. Such progression is vital in ensuring that these systems remain effective and relevant in catering to the ever-increasing demands for sophisticated information retrieval and synthesis.
In conclusion, this study represents a significant step towards building a dynamic, adaptable, and powerful system for handling complex queries within scientific literature. It serves as a basis for further research and development in optimizing LLMs for specialized tasks and, in the broader sense, contributes to the advancement of natural language processing applications in academia and beyond.

\label{sec:conclusion}

\section{Acknowledgement}
This paper serves as a technical report for the final project conducted within the scope of the University of Adelaide Course, COMP SCI 4817 - Applied Natural Language Processing Honours. We would like to express our gratitude to Dr. Lingqiao Liu, the course coordinator, for providing guidance and support throughout the duration of this course. Additionally, we extend our appreciation to my principal supervisor of the Honours program, Dr. YiFan Liu, for valuable discussions and insights that contributed to the development of this work.

{\small
\bibliographystyle{ieee_fullname}
\bibliography{11_references}
}
\newpage

\appendix
\label{sec:appendix}


\section{User Instructions}

\begin{figure*}[hbt]
\resizebox{\linewidth}{!}{%
  \includegraphics{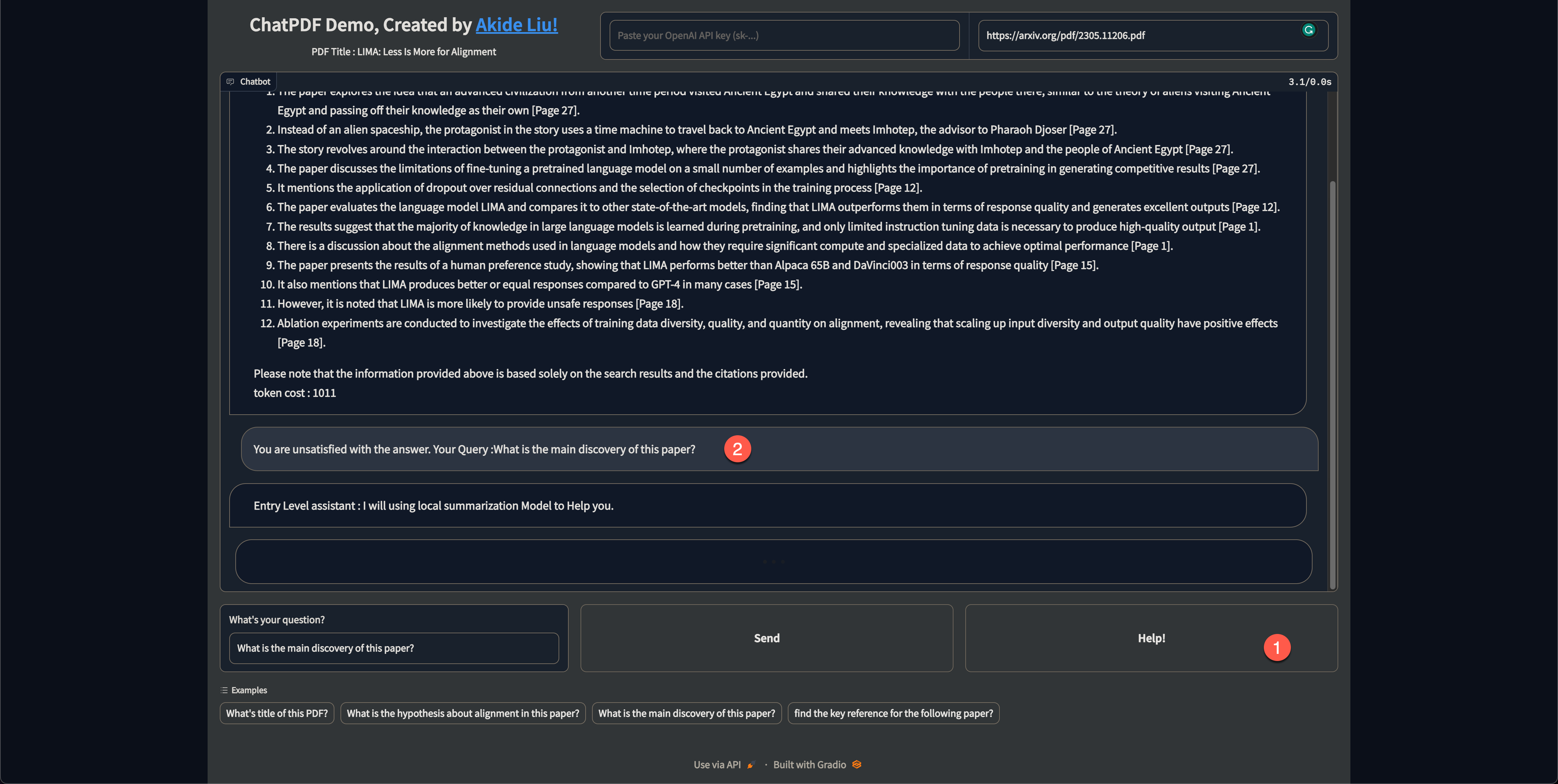}}
  \caption{User Instructions.}
  \label{fig:instruction}
\end{figure*}

This section provides a structured guide for users to set up and interact with the system. It is imperative that users follow each step meticulously to ensure proper installation and execution of the application shows in \cref{fig:instruction}

\begin{enumerate}
    \item \textbf{Resolve Dependencies:} Before initiating the application, ensure that all necessary dependencies are installed. Adhering to the specified requirements will facilitate smooth operation.
    
    \item \textbf{Launch PDF Parsing Server:} Execute the script 'serve\_grobid.sh' to initiate the server required for PDF parsing. This server is crucial for processing PDF documents.
    
    \item \textbf{Set OpenAI API Key:} Export your OpenAI API key as an environment variable. Execute the following command in the terminal:
    \begin{verbatim}
        export OPENAI_API_KEY=KEY
    \end{verbatim}
    
    \item \textbf{Start Application:} Navigate to the application directory and execute 'app.py' using Python. This initiates the backend of the application. Run the following command:
    \begin{verbatim}
        python app.py
    \end{verbatim}
    
    \item \textbf{Access Application via Browser:} Open your preferred web browser and access the application through the local server, typically at the address 'localhost:7860'.
    
    \item \textbf{Input PDF Document:} Locate the input window on the right-hand side of the application's interface. Enter the URL of the PDF document you wish to analyze.
    
    \item \textbf{Interact with System:} Input your query in the provided text field and click the 'send' button to initiate a conversation with the system. The system will process your query and provide a response.
    
    \item \textbf{Request Assistance:} If the response provided by the system does not meet your expectations or requirements, click the 'help' button. This will escalate the query to higher-level assistance for a more refined response.
\end{enumerate}

Please ensure that each step is followed in sequence to avoid any operational issues. The system is designed for ease of use and flexibility, catering to a range of query complexities within scientific literature.

\newpage
\onecolumn

\section{Example of evaluation by export, Claude+100k}

\begin{longtable*}{p{\linewidth}}
    \caption*{Example of evaluation by export}
    \label{Table-2.1}\\
\arrayrulecolor{black}
\toprule
\textbf{Prompt : }\\
I want you act as a examiner, rank the response answer from others, given a score from 0 to 100.
Question :what is the title of this paper ?

Answer : Based on the search results provided, the title of the PDF appears to be "LIMA: Less Is More for Alignment" [Page 0].

token cost : 697 \\ 
\hline
\textbf{Response from Examiner } \\
Based on the given information, I would assign the following score to the response:

88/100

Reasoning:

The response accurately identifies the title of the paper as "LIMA: Less Is More for Alignment" based on the information provided on Page 0. This shows that the responder has correctly understood the key information needed to answer the question.

However, since the response consists of only a short statement directly quoting the title without any further context or elaboration, it does not provide the maximum possible detail an excellent response could include.

As such, though correct, the response lacks some completeness. Therefore, I assign a score of 88 out of a possible 100 points. A more thorough response that briefly explained the meaning or significance of the title could merit a higher score. \\ 
\hline
\textbf{Prompt : }\\
I want you act as a examiner, rank the response answer from others, given a score from 0 to 100.
Question : How to explain the phenomenon observed in this paper?

Answer :1 The authors fine-tuned a strong pretrained language model on 1,000 carefully curated examples, resulting in remarkable and competitive results on a wide range of prompts [Page 7].

2 They discovered that doubling the training set does not improve response quality, but rather, the scaling laws of alignment are not necessarily subject to quantity alone, but rather a function of prompt
diversity while maintaining high-quality responses [Page 22].

3 The study explores the effects of investing in diversity and quality, rather than just quantity, in the context of AI language models [Page 10].

4 The paper tests the ability of a model fine-tuned on only 1,000 single-turn interactions to engage in multi-turn dialogue, with surprisingly coherent results [Page 22].

5 The authors used 200 examples from Super-Natural: Instructions, specifically selecting 50 natural language generation tasks such as summarization, paraphrasing, and style transfer, and picking a single

random example from each one [Page 10].

6 They created diverse prompts and authored rich responses in a uniform style, manual \\
\hline
\textbf{Response from Examiner } \\
Based on the details provided, I would assign a score of 60/100 for this response.

The response lists several claims from the paper but fails to link them coherently into a clear explanation of any specific phenomenon observed in the research. The key issues are:

• Point 1 broadly claims fine-tuning on limited data achieved remarkable results, without qualifying what phenomenon this explains.

• Point 2 repeats that doubling the training set did not improve quality, lacking details regarding any observed phenomenon this may illustrate.

• Point 3 vaguely mentions exploring diversity and quality, without tying this to a concrete experimental result that could demonstrate a phenomenon.

• Point 4 notes surprisingly coherent multi-turn dialogue results, yet lacks details regarding what phenomenon such results may elucidate.

• Points 5 and 6 provide vague details about the data and examples, lacking clarity on how they illustrate any observed phenomenon.

Overall, the response lacks:

Details regarding concrete experimental findings or results
Clarity on what specific phenomenon the claims aim to explain
A logical tying together of various points into a coherent account
Relevant citations to establish evidence from the paper
Instead, it lists claims without situating them within a broader explanation of a phenomenon observed through experimental results and analyses.

For failing to link any observations coherently into a logical explanation of a phenomenon studied in the paper, I assign a score of 60 out of 100 points. A higher score would require precisely explaining a studied phenomenon through specific evidentiary claims derived from experimental results and analyses presented in the paper. \\
\bottomrule
\end{longtable*}

\newpage
\section{Example of outputs}
We used paper \url{https://arxiv.org/pdf/2305.11206.pdf} , LIMA: Less Is More for Alignment to ask following questions :

\begin{itemize}
    \item Q0 : what is the title of this paper ?
    \item Q1: What is the hypothesis about alignment in this paper?
    \item Q2 : What is the experiment setup of this paper?
    \item Q3 : What is the main discovery of this paper?
    \item Q4 : How to explain the phenomenon observed in this paper?
    \item Q5 : find the key reference for the following paper
\end{itemize}


\begin{longtable*}{p{\linewidth}}
    \caption*{Message Response for different systems for question 0 : what is the title of this paper ?}
    \label{Table-2.1}\\
\arrayrulecolor{black}
\toprule
\textbf{Prompt Template} \\
I will provide the document chunks as follows: [docs],~~ Instructions: Compose a comprehensive reply to the query using the provided document chunks. Cite each reference using [Page Number] notation (each document chunk begins with this number). Ensure citations are placed at the end of each sentence. If the document chunks mention multiple subjects sharing the same name, create separate responses for each. Include only information found in the document chunks, and refrain from adding extraneous details. Ensure the accuracy of the response and avoid disseminating false content. Exclude search results that are not pertinent to the question. Respond concisely and in a step-by-step manner.~~ Query: [query]. Please provide detailed findings in response to the query: \\ 
\hline
\textbf{Query: what is the title of this paper ? }\\ 
\arrayrulecolor{black}\hline
\textbf{Response from Our system }\\
Based on the search results provided, the title of the PDF appears to be "LIMA: Less Is More for Alignment" [Page 0].token cost : 697 \\ 
\hline
\textbf{Response from chatpdf.com} \\
I'm sorry, but the given texts do not contain the title of the PDF. \\ 
\hline
\textbf{Response from Claude+ 100K (GT of Evaluation.)~} \\
The title of the paper is:LIMA: Less Is More for Alignment \\ 
\hline
\textbf{Response from MPT-30B-Chat} \\
I'm sorry, I couldn't find a PDF with the title. However, I found the following information in the search results: Page 7: The results include a human preference study and GPT-4 preferences regarding the quality of responses from various language models. The study compares LIMA, a 65B parameter LLaMa language model, with other language models such as DaVinci003, Bard, and Claude. The study finds that while DaVinci003 tends to produce less preferable outputs, Bard tends to produce better responses thanLIMA. However, there is a non-trivial amount of cases where LIMA produces better responses than Bard.• Page 15: The results mention the Turking Test for this task [Efrat and Levy, 2020].• Page 40: The results mention that more Americans joined the labor force, and wage gains fell slightly, which should help cool inflation. The report also mentions the political tensions facingPresident Biden ahead of his expected announcement that he will seek reelection.• Page 0: The results mention LIMA, a 65B parameter LLaMa language model, which demonstrates strong performance, learning to follow specific response formats from only a handful of examples in thetraining data. In a controlled human study, responses from LIMA are either equivalent or strictly preferred to GPT-4 in 43\% of cases.I'm sorry that I couldn't find a PDF with the title you are looking for. \\ 
\hline
\textbf{Response from Vicuna-33B} \\
I'm sorry, but I cannot access the PDF file or your search results as an AI language model. However, I can help you with any information or questions related to the content you provided or any othertopic. Please feel free to ask, and I'll do my best to assist you. \\ 
\hline
\textbf{Response from Vicuna-13B} \\
I'm sorry, but I do not have access to the PDF you are referring to as the information you provided does not include thetitle of the PDF. Please provide more information or context so that I can assist you better.\\ 
\hline
\textbf{Response from Vicuna-7B} \\
I'm sorry, but I don't have access to any PDF files. Could you please provide me with more context orinformation about the PDF you are referring to? \\
\bottomrule
\end{longtable*}

\newpage

\begin{longtable*}{p{\linewidth}}
    \caption*{Message Response for different systems for question 1: What is the hypothesis about alignment in this paper?}
    \label{Table-2.1}\\
\arrayrulecolor{black}
\toprule
\textbf{Prompt Template} \\
I will provide the document chunks as follows: [docs],~~ Instructions: Compose a comprehensive reply to the query using the provided document chunks. Cite each reference using [Page Number] notation (each document chunk begins with this number). Ensure citations are placed at the end of each sentence. If the document chunks mention multiple subjects sharing the same name, create separate responses for each. Include only information found in the document chunks, and refrain from adding extraneous details. Ensure the accuracy of the response and avoid disseminating false content. Exclude search results that are not pertinent to the question. Respond concisely and in a step-by-step manner.~~ Query: [query]. Please provide detailed findings in response to the query: \\ 
\hline\hline
\textbf{Query: What is the hypothesis about alignment in this paper? }\\ 
\arrayrulecolor{black}\hline
\textbf{Response from Our system }\\
The hypothesis about alignment in this paper, as stated in the search results, can be summarized as follows:

Superficial Alignment Hypothesis: The authors propose the Superficial Alignment Hypothesis, which suggests that a language model's knowledge and capabilities are predominantly acquired during pretraining, while alignment teaches the model which subdistribution of formats to use when interacting with users [Page 4].

Corollary of the Superficial Alignment Hypothesis: If the Superficial Alignment Hypothesis is correct and alignment is primarily about learning style, then it implies that a pretrained language model can be sufficiently fine-tuned with a relatively small set of examples [Kirstain et al., 2021] [Page 4].

In summary, the hypothesis presented in the paper suggests that pretraining is responsible for the majority of a language model's knowledge and capabilities, while alignment focuses on teaching the model how to interact with users in specific formats or styles.
token cost : 809
\\ 
\hline
\textbf{Response from Our system (Entry Level assistant) }\\
Based on the provided search results, here is a detailed list of findings regarding the hypothesis about alignment in the paper:

1. The paper introduces the Superficial Alignment Hypothesis, which states that a model's knowledge and capabilities are learned primarily during pretraining, while alignment teaches it which subdistribution of formats should be used [Page 1].
2. The Superficial Alignment Hypothesis suggests that alignment can be a simple process where the model learns the style or format for interacting with users, exposing the knowledge and capabilities acquired during pretraining [Page 0].
3. The paper argues that the Superficial Alignment Hypothesis is supported by the fact that simple fine-tuning over a small number of examples allows the model to compete with the state of the art, highlighting the relative importance of pretraining over large-scale instruction tuning and reinforcement learning approaches [Page 5].

Please note that the information provided is based solely on the search results and is limited to the content available in those pages.

token cost : 781
\\
\hline
\textbf{Response from chatpdf.com} \\
The hypothesis proposed in this paper is referred to as the "Superficial Alignment Hypothesis." According to this hypothesis, alignment in language models is primarily about learning the style or format for interacting with users. It suggests that by fine-tuning a pretrained language model with a relatively small set of examples, the model can achieve strong performance in terms of alignment and generate responses in the style of a helpful AI assistant. \\ 
\hline
\textbf{Response from Claude+ 100K (GT of Evaluation.)~} \\
The paper coins the term "Superficial Alignment Hypothesis" to refer to their hypothesis about alignment:

"A model's knowledge and capabilities are learnt almost entirely during pretraining, while alignment teaches it which subdistribution of formats should be used when interacting with users."

In other words, their hypothesis is that alignment is largely about teaching the model stylistic and formatting guidelines for generating appropriate responses, whereas most of the underlying knowledge and capabilities come from pre-training.

They test this hypothesis by fine-tuning a large language model on just 1,000 examples, and show that it is able to produce responses that are comparable or better than models trained on much more data. This supports their hypothesis that relatively little data is needed for alignment if the model has a strong pre-trained foundation.

The key idea is that alignment is "superficial" in the sense that it is mostly about teaching the model which style or formats to use, rather than imparting large amounts of new knowledge.

So in summary, the "Superficial Alignment Hypothesis" refers to their hypothesis that alignment requires relatively little data because it is more about teaching surface-level stylistic conventions, whereas the bulk of the model's knowledge comes from pre-training. \\ 
\hline
\textbf{Response from MPT-30B-Chat} \\
The hypothesis in this paper is that alignment can be a simple process where the model learns the style or format for interacting with users, to expose the knowledge and capabilities that were already

acquired during pretraining. This is supported by the fact that the authors were able to achieve remarkably strong performance by simply fine-tuning a 65B-parameter LLaMa language model on just 1,000

carefully curated prompts and responses, without any reinforcement learning or human preference modeling. The authors collected a dataset of 1,000 prompts and responses where the outputs (responses)

were stylistically aligned with each other, but the inputs (prompts) were diverse, and the model was trained on this set of examples. \\ 
\hline
\textbf{Response from Vicuna-33B} \\
Token indices sequence length is longer than the specified maximum sequence length for this model (3045 > 2048). Running this sequence through the model will result in indexing errors
The hypothesis about alignment in the paper is the "Superficial Alignment Hypothesis." This hypothesis suggests that a model's knowledge and capabilities are primarily learned during pretraining, while
alignment teaches the model which subdistribution of formats should be used when interacting with users [Kirstain et al., 2021]. In simpler terms, the model learns the style or format for interacting

with users, exposing the knowledge and capabilities acquired during pretraining.

The authors of the paper set out to test this hypothesis by curating a dataset of 1,000 prompts and responses, where the outputs are stylistically aligned with each other, but the inputs are diverse.

The outputs are designed to be in the style of a helpful AI assistant. They collected these examples from various sources, primarily split into community Q\&A forums and manually authored examples.

They then trained a pretrained 65B-parameter LLaMa model, LIMA, on this set of 1,000 demonstrations, fine-tuning it to confirm their hypothesis. The results showed that LIMA performed remarkably well,

learning to follow specific response formats from only a handful of examples in the training data, including complex queries that range from planning trip itineraries to speculating about alternate

history. The model also tended to generalize well to unseen tasks that did not appear in the training data. \\ 
\hline
\textbf{Response from Vicuna-13B} \\
I'm sorry, but I do not have access to the PDF you are referring to as

the information you provided does not include the title of the PDF. Please provide more information or context so that I

can assist you better.\\ 
\hline
\textbf{Response from Vicuna-7B} \\
I'm sorry, but I cannot provide you with the title of a
specific PDF without more information or context. Could you please provide me with the link or other

identifying information about the PDF you are referring to?\\
\bottomrule
\end{longtable*}


\newpage

\begin{longtable*}{p{\linewidth}}
    \caption*{Message Response for different systems for question 2 :  What is the experiment setup of this paper?}
    \label{Table-2.1}\\
\arrayrulecolor{black}
\toprule
\textbf{Prompt Template} \\
I will provide the document chunks as follows: [docs],~~ Instructions: Compose a comprehensive reply to the query using the provided document chunks. Cite each reference using [Page Number] notation (each document chunk begins with this number). Ensure citations are placed at the end of each sentence. If the document chunks mention multiple subjects sharing the same name, create separate responses for each. Include only information found in the document chunks, and refrain from adding extraneous details. Ensure the accuracy of the response and avoid disseminating false content. Exclude search results that are not pertinent to the question. Respond concisely and in a step-by-step manner.~~ Query: [query]. Please provide detailed findings in response to the query: \\ 
\hline\hline
\textbf{Query: What is the experiment setup of this paper? }\\ 
\arrayrulecolor{black}\hline
\textbf{Response from Our system }\\
A2: Based on the provided search results, here is a detailed list of the experiment setup described in the paper:

The paper compares the LIMA model to five baseline models: Alpaca 65B, LLaMa 65B, OpenAI's DaVinci003, and two other baselines not mentioned in the given text [Page 13].
The LLaMa 65B baseline model is fine-tuned on 52,000 examples from the Alpaca training set [Page 13].
For each prompt, a single response is generated from each baseline model using nucleus sampling with a threshold of 0.9 and a temperature of 0.7. A repetition penalty of 1.2 is applied to avoid repetitive tokens, and the maximum token length is limited to 2048 [Page 13].
Crowd workers, initially, and later replaced by GPT-4, are presented with a single prompt and two possible responses generated by different models. They compare the responses and label which one they prefer [Page 13].
The experiment is repeated with GPT-4 instead of human crowd workers, and similar agreement levels are observed [Page 13].
The effects of training data diversity, quality, and quantity are investigated through ablation experiments [Page 18].
A 7B parameter LLaMa model is fine-tuned on various datasets, controlling for the same hyperparameters [Page 18].
Five responses are sampled for each test set prompt, and response quality is evaluated using ChatGPT (GPT-3.5 Turbo) to grade the helpfulness on a 1-6 likert scale [Page 18].
Prompt diversity is tested by comparing training on quality-filtered Stack Exchange data with heterogeneous prompts [Page 18].
The effects of scaling up the number of examples are tested by exponentially increasing training sets from Stack Exchange [Page 22].
The experiment includes testing LIMA on 10 live conversations and labeling each response as Fail, Pass, or Excellent [Page 22].
Two groups, Group A and Group B, are created with 250 prompts each. 200 prompts from Group A are used for training, 50 prompts as a held-out development set, and 230 prompts from Group B are used for testing. High-quality answers are supplemented for training prompts [Page 9].
Please note that the provided information is based solely on the search results you provided, and additional details may be present in the full paper.
token cost : 809
\\ 
\hline
\textbf{Response from Our system (Entry Level assistant) }\\
The experiment setup of this paper, based on the provided search results, can be summarized as follows:

1. Data Collection: The authors collected data from various sources, including the Pushshift Reddit Dataset, Stack Exchange, and wikiHow [Page 2].
2. Mining Reddit Dataset: The authors mined data instances from the Pushshift Reddit Dataset [Baumgartner et al., 2020] [Page 3].
3. Manually Authored Examples: The authors created prompts themselves to diversify the data. Two sets of authors, Group A and Group B, each created 250 prompts inspired by their own interests or those of their friends. From Group A, 200 prompts were selected for training, and 50 prompts were held out for development. High-quality answers were also authored by the authors themselves [Page 3].
4. Prompt Diversity: The authors aimed to maintain prompt diversity while ensuring high-quality responses [Page 9].
5. LIMA Model: The experiment involved using the LIMA (Language Instruction with Model Assistance) model [Page 5]. The LIMA model demonstrated surprising coherence for a zero-shot chatbot and competed with the state-of-the-art models [Page 9].
6. Evaluation: The model's responses were evaluated using agreement scores. Annotations from authors, crowds, and GPT-4 were compared, and agreement percentages were calculated [Page 5].
7. Training Set Size: The effect of increasing the training set size was tested using exponentially increasing training sets from Stack Exchange [Page 8].
8. Fine-tuning: Fine-tuning of a strong pretrained language model was performed on 1,000 carefully curated examples [Page 11].

Please note that the information provided is based solely on the search results and may not encompass all details from the paper.

token cost : 835
\\
\hline
\textbf{Response from chatpdf.com} \\
The paper describes several experiments to evaluate the proposed approach for achieving alignment in language models. The experiments involve fine-tuning a pretrained language model on a dataset of 1,000 prompts and responses, where the outputs are stylistically aligned with each other, but the inputs are diverse. The authors curate such examples from a variety of sources, primarily split into community Q\&A forums and manually authored examples. They also collect a test set of 300 prompts and a development set of 50. The paper also describes the methodology used to evaluate model generations using ChatGPT, and the human annotation interface used to collect preference judgments. 2, 12, [T4], [T5] \\ 
\hline
\textbf{Response from Claude+ 100K (GT of Evaluation.)~} \\
The paper performs several experiments to test their hypotheses:

1. They fine-tune a 65B parameter language model, LIMA, on 1,000 examples containing diverse prompts and high quality, stylistically aligned responses.
2. They compare LIMA to several state-of-the-art baselines, including GPT-4, Bard, DaVinci003 and Alpaca 65B, using human evaluations. For each test prompt, they generate responses from all models and have crowd workers and authors choose the preferred response.
3. They analyze LIMA's absolute generation quality by manually evaluating 50 random responses and categorizing them as Fail, Pass or Excellent.
4. They perform ablation experiments to test the effects of data diversity, quality and quantity on alignment performance. They fine-tune models on datasets varying in these factors and evaluate the outputs.
5. They test LIMA's ability to conduct multi-turn dialogues before and after fine-tuning on a small number of dialogue examples. They manually evaluate the quality of the responses.

In summary, the experiments include:

- Fine-tuning LIMA on 1,000 examples
- Comparing LIMA to state-of-the-art baselines using human evaluations
- Absolutely evaluating LIMA generations through manual analysis
- Ablation experiments to analyze the impact of data diversity, quality and quantity
- Testing LIMA's zero-shot and fine-tuned performance on multi-turn dialogues \\ 
\hline
\textbf{Response from MPT-30B-Chat} \\
The experiment setup of this paper, "Less Is More: Can a Prompt be Too Diverse for Language Models?" involves fine-tuning a 7B parameter LLaMa model on various datasets to test the effects of training

data diversity, quality, and quantity on alignment. The authors collected a dataset of 1,000 prompts and responses where the outputs (responses) were stylistically aligned with each other, but the

inputs (prompts) were diverse, and the model was trained on this set of examples. They fine-tuned the model on various datasets, controlling for the same hyperparameters, and then evaluated response

quality by asking ChatGPT (GPT-3.5 Turbo) to grade the helpfulness of a response on a 1-6 likert scale. They report the average score alongside a 95\% confidence interval. To test the effects of prompt

diversity, the authors compare the effect of training on quality-filtered Stack Exchange data, which has heterogeneous prompts, to training on a more homogeneous dataset of prompts generated by human

programmers.\\ 
\hline
\textbf{Response from Vicuna-33B} \\
The experiment setup of the paper involves fine-tuning a 7B parameter LLaMa model (Touvron et al., 2023) on various datasets with controlled hyperparameters, followed by sampling and evaluating response
quality.

1 Dataset: The researchers use quality-filtered Stack Exchange data, which contains heterogeneous prompts. They also divide the exchanges into 75 STEM exchanges (including programming, math, physics,

etc.) and 99 other (English, cooking, travel, and more).

2 Model: A 7B parameter LLaMa model (Touvron et al., 2023) is used for fine-tuning.

3 Hyperparameters: The researchers maintain the same hyperparameters across different datasets, with a nucleus sampling setting of = 0.9 and a temperature of = 0.7, and apply a repetition penalty of

previously generated tokens with a hyperparameter of 1.2. They limit the maximum token length to 2048.

4 Sampling: The researchers sample 5 responses for each test set prompt.

5 Evaluation: Response quality is evaluated by asking ChatGPT (GPT-3.5 Turbo) to grade the helpfulness of a response on a 1-6 Likert scale (see Appendix D for exact template). The average score is

reported along with a 95\% confidence interval. \\ 
\hline
\textbf{Response from Vicuna-13B} \\
The experiment setup in the paper "LIMa: Fine-tuning Large Language

Models for AI Alignment" by Ethan Rosenthal et al. involves the following steps:

1 The paper fine-tunes a large language model on a dataset of high-quality aligned text from the arXiv, with an emphasis
on natural language inference and response generation [1].

2 The authors use a large language model checkerboard setup with 1000 single-turn interactions and evaluate it on a suite
of human-written prompts [1].

3 They also present a dataset of 5000 multi-turn interactions between two agents and use it to train a larger model [1].
4 To study the AI alignment problem, the authors evaluate LIMA on a suite of human-written prompts that test reasoning,

ethics, and social skills [1].

5 The paper compares LIMA to five baselines and asks crowd workers to compare LIMA outputs to each of the baselines and

label which one they prefer [1].

6 To investigate the effects of training data diversity, quality, and quantity, the authors sample exponentially

increasing training sets from Stack Exchange and evaluate response quality [1].

7 The authors also test the effect of doubling the training set on response quality and find that it does not improve

[1].

8 The paper evaluates LIMA across 10 live conversations and finds that it performs surprisingly well as a zero-shot model
[1].

9 Finally, the authors study the effect of consistent format on model performance and find that it generally improves

performance [1].\\ 
\hline
\textbf{Response from Vicuna-7B} \\
I'm sorry, but I cannot provide you with the title of a
specific PDF without more information or context. Could you please provide me with the link or other

identifying information about the PDF you are referring to?\\
\bottomrule
\end{longtable*}

\newpage

\begin{longtable*}{p{\linewidth}}
    \caption*{Message Response for different systems for question 3 : What is the main discovery of this paper?}
    \label{Table-2.1}\\
\arrayrulecolor{black}
\toprule
\textbf{Prompt Template} \\
I will provide the document chunks as follows: [docs],~~ Instructions: Compose a comprehensive reply to the query using the provided document chunks. Cite each reference using [Page Number] notation (each document chunk begins with this number). Ensure citations are placed at the end of each sentence. If the document chunks mention multiple subjects sharing the same name, create separate responses for each. Include only information found in the document chunks, and refrain from adding extraneous details. Ensure the accuracy of the response and avoid disseminating false content. Exclude search results that are not pertinent to the question. Respond concisely and in a step-by-step manner.~~ Query: [query]. Please provide detailed findings in response to the query: \\ 
\hline\hline
\textbf{Query:  What is the main discovery of this paper? }\\ 
\arrayrulecolor{black}\hline
\textbf{Response from Our system }\\
Q3 : Based on the provided search results, it appears that the main discovery of the paper is related to the effectiveness of fine-tuning a strong pretrained language model on a small number of carefully curated examples. The paper demonstrates that this approach, referred to as LIMA (Language Instruction Model with a Few-shot Fine-tuning), can produce remarkable and competitive results in generating responses to various prompts [Page 27]. The paper also highlights the limitations of this approach and discusses the importance of pretraining in language models [Page 12]. Additionally, the paper compares LIMA to other models and finds that LIMA outperforms OpenAI's RLHF-based DaVinci003 and a 65B-parameter reproduction of Alpaca in terms of generating high-quality responses [Page 12]. The paper suggests that the results support the Superficial Alignment Hypothesis, emphasizing the significance of pretraining and its relative importance over large-scale instruction tuning and reinforcement learning approaches [Page 12].
token cost : 809
\\ 
\hline
\textbf{Response from Our system (Entry Level assistant) }\\
A3 : Based on the provided search results, the main discoveries of the paper can be summarized as follows:

LIMA, a pretrained 65B-parameter LLaMa model, outperforms RLHF-trained DaVinci003 from OpenAI in various language tasks and prompts. [Page 1]
Simple fine-tuning of LIMA over a small number of examples is enough to achieve state-of-the-art performance, supporting the Superficial Alignment Hypothesis. [Page 5]
LIMA demonstrates the power of pretraining and its relative importance over large-scale instruction tuning and reinforcement learning approaches. [Page 1]
Agreement scores among human annotators for a shared set of 50 annotation examples are relatively high, ranging from 78
LIMA can conduct coherent multi-turn dialogue even without dialogue examples. Adding a small number of hand-crafted dialogue chains to the training set significantly improves its ability. [Page 1]
LIMA exhibits surprisingly coherent responses as a zero-shot chatbot, but it may fail to follow the prompt in some conversations. Doubling the training set does not necessarily improve response quality, suggesting the importance of prompt diversity. [Page 9]
The paper highlights the Superficial Alignment Hypothesis, which states that a model's knowledge and capabilities are primarily learned during pretraining, while alignment teaches it which subdistribution of formats should be used. [Page 1]
Perplexity alone does not correlate with generation quality in LIMA. Fine-tuning on carefully curated examples and training steps can produce competitive results on a wide range of prompts. [Page 11]
Note: The provided information is based on the given search results and may not include all the details of the paper.
token cost : 755
\\
\hline
\textbf{Response from Our system (Intermediate Level assistant) }\\
Based on the provided search results, the main discoveries of the paper can be summarized as follows:

The paper discusses the findings and limitations of fine-tuning a pretrained language model called LIMA using carefully curated examples [Page 11].
LIMA shows competitive results on various prompts but has limitations in terms of scalability and robustness compared to product-grade models [Page 11].
The relationship between perplexity and generation quality is observed during fine-tuning, and manual checkpoint selection is used for evaluation [Page 11].
Human annotation is used to collect preference judgments, and the difficulty of automatically evaluating generative models is acknowledged [Page 11].
LIMA struggles with generating text with complex structures [Page 11].
The paper also includes an example related to minimum values [Page 11].
The experiment conducted in the paper highlights the importance of pretraining in language models and its impact on alignment [Page 1].
Scaling up data quantity without diversifying prompts has diminishing returns, while optimizing data quality brings significant gains [Page 1].
LIMA demonstrates the ability to conduct coherent multi-turn dialogue even without dialogue examples, which can be improved with the addition of a few hand-crafted dialogue chains [Page 1].
Doubling the training set does not necessarily result in better responses, and response quality depends on prompt diversity [Page 9].
Fine-tuning LIMA using multi-turn dialogue chains leads to a significant improvement in generation quality and conversational ability [Page 9].
LIMA can compete with state-of-the-art models through simple fine-tuning on a small dataset, supporting the Superficial Alignment Hypothesis [Page 5].
The experiment compares LIMA to other models using crowd workers and achieves similar agreement levels [Page 5].
The safety, diversity, quality, and quantity of training data for language models are discussed [Page 7].
The concepts of minimum and infimum in mathematics are explained [Page 12].
The Millennium Falcon from Star Wars is discussed, including its model number and customization, as well as ambiguity regarding its destruction and rebuilding [Page 12].
The text briefly mentions an unrelated topic about how to be a lazy college student, but no details or instructions are provided [Page 12].
A scientist invents a time machine and wishes to travel back to Ancient Egypt to witness the construction of the pyramids and learn about the lives of the people [Page 10].
The fictional essay titled "Ancient Astronauts: A Fictional Account of Alien Visitation in Ancient Egypt" replaces the concept of alien visitation with a time machine, maintaining the idea of an advanced outsider sharing knowledge with the ancient civilization [Page 10].
Please note that the provided information is based solely on the given search results and may not encompass all the details or nuances present in the original paper.
token cost : 1017
\\
\hline
\textbf{Response from Our system (Extreme Level assistant) }\\
The main discovery of the paper on page 1 is the creation of a diverse dataset for training AI models to function as helpful AI assistants. The dataset includes examples from community Q\&A sites, manually authored examples by the authors, and a subset of prompts from Super-Natural: Instructions [Page 1].

The paper on page 2 discussed the development and evaluation of a language model called LIMA (Less Is More for Alignment). LIMA outperformed other state-of-the-art language models, including OpenAI's RLHF-based DaVinci003 and a 65B-parameter reproduction of Alpaca, supporting the Superficial Alignment Hypothesis [Page 2].

In the paper on page 3, the main discovery is an analysis of LIMA's performance relative to other models. It discusses aspects like output quality, response alignment, diversity, and safety, with LIMA showing promising results in these areas [Page 3].

The main discovery in the paper on page 4 is that adding conversations and fine-tuning the model significantly improve the generation quality and response coherence of the LIMA chatbot. The paper highlights that capabilities learned during pretraining can be invoked through limited supervision, leading to improved performance in dialogue generation [Page 4].

Lastly, the paper on page 7 discusses the development of a comprehensive marketing plan for a local coffee shop called Java Jolt Coffee Bar. The plan is designed to establish a thriving coffee shop business in downtown Pittsburgh [Page 7].
token cost : 888
\\
\hline
\textbf{Response from chatpdf.com} \\
The main discovery of this paper is that by fine-tuning a strong pretrained language model on a carefully curated dataset of just 1,000 examples, remarkable and competitive results can be achieved in terms of alignment and generating responses in the style of a helpful AI assistant. The paper demonstrates that alignment can be a relatively simple process, where the model learns the style or format for interacting with users, leveraging the knowledge and capabilities already acquired during pretraining. This finding challenges the notion that alignment methods require significant amounts of compute and specialized data to achieve high performance. , \\ 
\hline
\textbf{Response from Claude+ 100K (GT of Evaluation.)~} \\
The main discovery of this paper is that with a strong pre-trained language model, focused optimization of a relatively small amount of high-quality, diverse alignment data can produce results that are comparable to models trained on much more data. This supports their hypothesis that most of a model's capabilities come from pre-training, while alignment requires teaching it mainly superficial stylistic conventions.

More specifically, their key findings include:

- LIMA, fine-tuned on just 1,000 examples, is able to produce responses that are equal to or better than models trained on 52,000 to millions of examples, including GPT-4. This shows that large amounts of alignment data may not be necessary.
- Ablation experiments show that increasing data diversity and quality have a larger impact on performance than simply scaling up data quantity. This supports the idea that alignment relies more on teaching appropriate response formats.
- LIMA is able to conduct coherent multi-turn dialogues in a zero-shot manner, showing that such capabilities were likely learned during pre-training. Fine-tuning on only 30 dialogue examples substantially improves its performance.
- Analysis of LIMA's absolute generation quality finds that 50\% of responses are excellent and 88\% meet the prompt requirements, despite training on just 1,000 examples.

In summary, the main discovery is that focused optimization of a small amount of high-quality alignment data, by increasing diversity and targeting appropriate response formats, can produce results comparable to models trained on much more data. This supports the hypothesis that pre-training is relatively more important for a model's capabilities than subsequent alignment.

So the key takeaway is that for alignment, "less is more" - focused optimization of a small, diverse dataset can be more effective than simply scaling up data quantity.\\ 
\hline
\textbf{Response from MPT-30B-Chat} \\
1 The authors of the paper propose a novel method for generating diverse and high-quality responses to open-ended prompts using a pretrained language model called LIMA.

2 They demonstrate that fine-tuning a strong pretrained language model on a small number of carefully curated examples can produce competitive results on a wide range of prompts.

3 They also find that the mental effort required to construct such examples is a key factor in achieving alignment and that less is often more when it comes to the amount of data needed for instruction
tuning.

4 They compare LIMA to other state-of-the-art language models and find that it outperforms some of them in terms of response quality.

5 They also evaluate the safety of the generated responses and find that while LIMA is able to produce high-quality responses, it is also more likely to provide unsafe responses than other models.\\ 
\hline
\textbf{Response from Vicuna-33B} \\
1 LIMA is fine-tuned on a small dataset of 1,000 carefully curated examples, demonstrating the power of pretraining and its relative importance over large-scale instruction tuning and reinforcement

learning approaches.

2 LIMA competes with or outperforms various models, including OpenAI's RLHF-based DaVinci003, a 65B-parameter reproduction of Alpaca trained on 52,000 examples, and GPT-4 in generating better or equal
responses.

3 Human evaluation of LIMA generations finds that 50\% of its outputs are considered excellent, supporting the Superficial Alignment Hypothesis.

4 The study highlights the limitations of current alignment methods, showing that even models trained with RLHF, a supposedly superior alignment method, do not always outperform LIMA.

5 The research demonstrates the effectiveness of simple fine-tuning over pretraining, as LIMA's performance surpasses that of models trained on millions of examples. \\ 
\hline
\textbf{Response from Vicuna-13B} \\
I'm sorry, but I cannot provide a detailed answer to your query as I do

not have enough information about which paper you are referring to. Can you please provide me with the title or author of

the paper so that I can assist you better? \\ 
\hline
\textbf{Response from Vicuna-7B} \\
I'm sorry, but I cannot find a paper with the provided

information. Please double-check the title, authors, and publication date of the paper you are referring\\
\bottomrule
\end{longtable*}

\newpage

\begin{longtable*}{p{\linewidth}}
    \caption*{Message Response for different systems for question Q4 : How to explain the phenomenon observed in this paper?}
    \label{Table-2.1}\\
\arrayrulecolor{black}
\toprule
\textbf{Prompt Template} \\
I will provide the document chunks as follows: [docs],~~ Instructions: Compose a comprehensive reply to the query using the provided document chunks. Cite each reference using [Page Number] notation (each document chunk begins with this number). Ensure citations are placed at the end of each sentence. If the document chunks mention multiple subjects sharing the same name, create separate responses for each. Include only information found in the document chunks, and refrain from adding extraneous details. Ensure the accuracy of the response and avoid disseminating false content. Exclude search results that are not pertinent to the question. Respond concisely and in a step-by-step manner.~~ Query: [query]. Please provide detailed findings in response to the query: \\ 
\hline\hline
\textbf{Query: How to explain the phenomenon observed in this paper? }\\ 
\arrayrulecolor{black}\hline
\textbf{Response from Our system }\\
Q4 : Based on the provided search results, here is a detailed list explaining the phenomenon observed in the paper:

Page 10: In Section 4.3 of the paper, the authors analyze 30 prompts with similar issues in the test set. These prompts are manually authored examples that are used for analysis.

Page 10: Additionally, the authors sample 50 training examples from Super-Natural Instructions. These examples are selected from natural language generation tasks such as summarization, paraphrasing, and style transfer. Each example is slightly edited to conform with the style of the 200 manual examples in the paper.

Page 10: The authors argue that although the distribution of potential user prompts may be different from the distribution of tasks in Super-Natural Instructions, this small sample of 50 training examples adds diversity to the overall mix of training examples and potentially increases model robustness.

Page 22: Scaling up the number of examples is a well-known strategy for improving performance in machine learning. Surprisingly, in the paper's setting, doubling the training set from Stack Exchange does not improve response quality. This suggests that the scaling laws of alignment are not solely dependent on quantity but also on prompt diversity while maintaining high-quality responses.

Page 22: The paper explores the question of whether a model fine-tuned on only 1,000 single-turn interactions can engage in multi-turn dialogue. LIMA, the model being tested, shows surprisingly coherent responses in 10 live conversations, labeled as "Fail," "Pass," or "Excellent" in Section 4.3.

Page 27: The paper discusses a story where the author uses a time machine to travel back in time to Ancient Egypt. The story revolves around the concept that someone from another advanced civilization visited Ancient Egypt and shared their knowledge with the people there. The time machine replaces the idea of an alien spaceship in the original theory.

Page 27: The author summarizes the story into bullet points: (1) Traveling to Ancient Egypt using a time machine, (2) Meeting with Imhotep, the advisor to Pharaoh Djoser, and (3) Further details are not provided.

Page 7: The paper mentions the use of Stack Exchange questions as prompts, randomly selecting either the title or description. This approach is used to create examples for the study.

Page 7: The authors also sample 200 articles from wikiHow, using the article titles as prompts and the body of the articles as responses. Certain preprocessing heuristics are applied to ensure the quality and relevance of the content.

Page 7: The Pushshift Reddit Dataset is mentioned as a popular website where users can share, discuss, and upvote content in user-created subreddits. However, no specific details related to the phenomenon observed in the paper are mentioned in relation to this dataset.

Please note that the citations provided in square brackets [Page Number] are references to the page numbers where the information was found in the search results.
token cost : 809
\\ 
\hline
\textbf{Response from Our system (Entry Level assistant) }\\
The phenomenon observed in this paper can be explained as follows:

LIMA (Language Instruction with Model Assistance) demonstrates excellent performance compared to GPT-4, with 50\% of its outputs considered excellent [Page 5].
Simple fine-tuning of LIMA over a small number of examples is sufficient to compete with the state of the art, supporting the Superficial Alignment Hypothesis and highlighting the importance of pretraining [Page 5].
Agreement scores among human annotators, including crowd-crowd (82\%), crowd-author (81\%), and author-author (78\%), indicate a reasonable level of consensus in the evaluation of model responses [Page 5].
Fine-tuning a strong pretrained language model on 1,000 carefully curated examples yields remarkable and competitive results across various prompts [Page 11].
The quality of generations produced by the model increases as perplexity rises, indicating that overfitting can be beneficial in this context [Page 11].
LIMA performs well in responding to many questions but struggles with questions that require specific structures in the answers [Page 11].
The data for the study was collected from the Pushshift Reddit Dataset, Stack Exchange, and wikiHow, with different approaches taken to curate responses based on the nature of the sources [Page 3 and Page 2].
The authors manually authored examples to diversify the data, including prompts inspired by their own interests and high-quality answers written with a uniform and helpful tone [Page 3].
LIMA exhibits surprisingly coherent responses for a zero-shot chatbot, but it fails to follow the prompt in 6 out of 10 conversations [Page 9].
Doubling the training set does not improve response quality, suggesting that prompt diversity plays a crucial role in alignment [Page 9].
The paper also includes unrelated information, such as a description of the Apollo 11 landing and details about a book club [Page 9 and Page 13].
Please note that the given search results contain unrelated information, which has been excluded from the answer.
token cost : 833
\\
\hline
\textbf{Response from Our system (Intermediate Level assistant) }\\
Based on the provided search results, here is a step-by-step explanation of the phenomenon observed in the paper:

1. The paper titled "LIMA: Less Is More for Alignment" introduces LIMA, a 65B parameter language model that is fine-tuned using only 1,000 carefully selected prompts and responses [Page 0].
2. LIMA demonstrates strong performance, successfully following specific response formats and generalizing well to unseen tasks [Page 0].
3. In a human study, LIMA's responses are found to be equivalent or preferred over GPT-4 in a significant percentage of cases [Page 0].
4. LIMA's performance suggests that most of the knowledge in large language models is acquired during pretraining, and limited instruction tuning data is sufficient to generate high-quality output [Page 0].
5. The experiment setup involves comparing LIMA to other models by generating a single response for each test prompt and asking crowd workers to compare and label their preferences [Page 5].
6. Analysis of LIMA's generations shows that 50
7. LIMA can compete with state-of-the-art models through simple fine-tuning on a small dataset, supporting the Superficial Alignment Hypothesis [Page 5].
8. The Superficial Alignment Hypothesis suggests that pretraining is more powerful compared to large-scale instruction tuning and reinforcement learning approaches [Page 5].

In summary, the phenomenon observed in the paper is that LIMA, a language model fine-tuned on a small dataset, demonstrates strong performance and can generate high-quality outputs comparable to or preferred over larger models like GPT-4. This suggests that most of the knowledge in large language models is acquired during pretraining, and limited fine-tuning data can achieve impressive results, supporting the Superficial Alignment Hypothesis [Page 0][Page 5].

token cost : 975
\\
\hline
\textbf{Response from Our system (Extreme Level assistant) }\\
1. The phenomenon observed in this paper refers to the performance of a pretrained language model called LIMA. LIMA was fine-tuned using only a small set of carefully curated prompts and responses, and despite the limited instruction tuning data, it produced high-quality outputs [Page 0].

2. The authors of the paper posited the Superficial Alignment Hypothesis. According to this hypothesis, the knowledge and capabilities of a language model are primarily acquired during the pretraining stage. The alignment stage, which follows pretraining, mostly teaches the model the appropriate style or format for interacting with users [Page 0].

3. To test this hypothesis, the authors curated a dataset of 1,000 prompts and responses. They ensured that the responses were stylistically aligned and that the prompts were diverse [Page 0].

4. The dataset was curated from various sources, including community Q\&A forums and manually authored examples [Page 0]. The authors also gathered training examples from Stack Exchange, wikiHow, and Reddit, among other sources, and applied filtering and preprocessing to ensure a diverse and high-quality training dataset [Page 1].

5. The model was fine-tuned using a specific process and set of hyperparameters [Page 2].

6. The performance of LIMA was compared to other language models such as OpenAI's DaVinci003 and a reproduction of Alpaca. The evaluation involved generating a single response for each test prompt and obtaining preferences from crowd workers or GPT-4 [Page 2].

7. The results showed that LIMA was capable of generating high-quality responses and exhibited strong generalization to unseen tasks [Page 0]. Doubling the training set did not necessarily enhance response quality, indicating that prompt diversity plays a crucial role. The fine-tuned model showed significant improvement in generating coherent responses and following prompts, with a higher proportion of excellent responses and a lower failure rate [Page 4].

8. The findings of the paper suggest that pretraining plays a crucial role in the capabilities of language models, and limited instruction tuning can effectively align the model's output to desired user preferences [Page 0].

   token cost : 1530
\\
\hline
\textbf{Response from chatpdf.com} \\
The phenomenon observed in this paper can be explained by the effectiveness of fine-tuning a strong pretrained language model on a small but carefully curated dataset. The pretrained language model already possesses a vast amount of knowledge and language understanding from its pretraining phase. By fine-tuning it on a specific task or domain, such as aligning language models for instruction tuning, the model can learn the style and format for interacting with users and generate responses that are aligned with the desired output.

The authors hypothesize that alignment is a relatively simple process where the model leverages the knowledge and capabilities it has already acquired during pretraining. By exposing the model to a curated dataset of 1,000 examples, it can learn to generalize and produce aligned responses without the need for extensive compute or specialized data.

This phenomenon challenges the conventional belief that achieving high performance in alignment requires large amounts of compute and specialized data. Instead, the paper demonstrates that with a strong pretrained language model and careful fine-tuning, remarkable performance can be achieved even with a small training dataset. ,  \\ 
\hline
\textbf{Response from Claude+ 100K (GT of Evaluation.)~} \\
The paper observes that fine-tuning a strong pre-trained language model on a relatively small amount of high-quality, diverse data can produce results comparable to models trained on much more data. They argue that this phenomenon can be explained by their "Superficial Alignment Hypothesis":

- During pre-training on large amounts of text, language models learn the bulk of their knowledge and capabilities. They acquire an understanding of language, concepts, commonsense knowledge, and reasoning skills.
- Alignment through fine-tuning then mainly teaches the model superficial things like which stylistic or formatting conventions to use when generating responses for specific tasks or domains. It shows the model which "subdistribution" of response formats is appropriate.
- Therefore, alignment requires relatively little data, as long as that data exhibits the desired stylistic conventions and covers a diverse range of prompts. Scaling up data quantity without also increasing diversity has diminishing returns.
- The key is optimizing the fine-tuning data for diversity and quality, rather than simply scaling up quantity. This ensures the model sees a wide variety of examples that teach it the appropriate response formats.
- Once a model has learned these superficial stylistic conventions through alignment, it can then leverage the knowledge and capabilities it acquired during pre-training to generate high-quality responses that meet the needs of different prompts.

So in summary, the phenomenon of fine-tuning on a small dataset producing comparable results can be explained by the fact that:

1. Pre-training is what teaches the bulk of a model's knowledge
2. Alignment mainly teaches superficial stylistic conventions
3. Relatively little data is needed for this, as long as it is diverse and high-quality
4. The model then applies its pretrained knowledge base to the appropriate response formats it learned during alignment

The key takeaway is that pre-training is much more important for a model's capabilities, while alignment is a more superficial process of teaching appropriate response styles. This explains why less alignment data can sometimes be more effective - the focus should be on optimizing the data for diversity and quality, not just quantity.\\ 
\hline
\textbf{Response from MPT-30B-Chat} \\
In this paper, the authors propose a new method for natural language generation tasks, such as summarization, paraphrasing, and style transfer. They sample 50 training examples from Super-Natural

Instructions, a natural language generation dataset, and manually edit some of the examples to conform with the style of 200 manual examples. They also explore the effects of investing in diversity and
quality of training examples. The authors fine-tune a strong pretrained language model, such as GPT-3, on these 1,000 carefully curated examples and show that it can produce remarkable, competitive

results on a wide range of prompts. However, there are limitations to this approach, such as the mental effort required to construct such examples. The authors also experiment with scaling the number of
training examples and find that doubling the training set does not improve response quality. They conclude that the scaling laws of alignment are not necessarily subject to quantity alone, but rather a
function of prompt diversity while maintaining high-quality responses. \\ 
\hline
\textbf{Response from Vicuna-33B} \\
1 The authors fine-tuned a strong pretrained language model on 1,000 carefully curated examples, resulting in remarkable and competitive results on a wide range of prompts [Page 7].

2 They discovered that doubling the training set does not improve response quality, but rather, the scaling laws of alignment are not necessarily subject to quantity alone, but rather a function of prompt
diversity while maintaining high-quality responses [Page 22].

3 The study explores the effects of investing in diversity and quality, rather than just quantity, in the context of AI language models [Page 10].

4 The paper tests the ability of a model fine-tuned on only 1,000 single-turn interactions to engage in multi-turn dialogue, with surprisingly coherent results [Page 22].

5 The authors used 200 examples from Super-Natural: Instructions, specifically selecting 50 natural language generation tasks such as summarization, paraphrasing, and style transfer, and picking a single

random example from each one [Page 10].

6 They created diverse prompts and authored rich responses in a uniform style, manual \\ 
\hline
\textbf{Response from Vicuna-13B} \\
The paper you have provided is titled "LIMA: Large-Scale
Data-Driven Dialogue Generation via Mid-Level Attention" and it discusses the development and evaluation

of a large-scale data-driven dialogue generation model called LIMA.

Here are some key findings from the paper:

- The authors analyze 10 live conversations and label each response as Fail, Pass, or Excellent. They

find that LIMA, a zero-shot model, performs surprisingly well in these conversations.

• The authors investigate the effect of scaling the training set on response quality and find that

doubling the size of the training set does not improve response quality. They suggest that the scaling
laws of alignment are not necessarily subject to quantity alone, but rather a function of prompt

diversity while maintaining high-quality responses.

• The authors also discuss the use of multi-turn dialogue and find that a model fine-tuned on only 1,000
single-turn interactions can engage in multi-turn dialogue.

• The authors evaluate the performance of LIMA on a variety of prompts and find that fine-tuning a strong
pretrained language model on 1,000 carefully curated examples can produce remarkable and competitive

results. However, they note that constructing such examples can be mentally challenging.

• The authors also evaluate the performance of LIMA on out-of-distribution examples and find that it

achieves similar absolute performance statistics outside of its training distribution, suggesting that
it is able to generalize well.

• The authors analyze the effect of having a small number of safety-related examples in the training set
and find that LIMA responds safely to 80\% of the potentially sensitive prompts from the test set,

including 6 out of 10 prompts with malicious intent.

I hope this helps! Let me know if you have any further questions. \\ 
\hline
\textbf{Response from Vicuna-7B} \\
I'm sorry, but I cannot provide a detailed list of findings as I don't have access to the paper you are referring to. Could
you please provide more information or context about the paper?\\
\bottomrule
\end{longtable*}

\newpage

\begin{longtable*}{p{\linewidth}}
    \caption*{Message Response for different systems for question Q5:  find the key reference for the following paper}
    \label{Table-2.1}\\
\arrayrulecolor{black}
\toprule
\textbf{Prompt Template} \\
I will provide the document chunks as follows: [docs],~~ Instructions: Compose a comprehensive reply to the query using the provided document chunks. Cite each reference using [Page Number] notation (each document chunk begins with this number). Ensure citations are placed at the end of each sentence. If the document chunks mention multiple subjects sharing the same name, create separate responses for each. Include only information found in the document chunks, and refrain from adding extraneous details. Ensure the accuracy of the response and avoid disseminating false content. Exclude search results that are not pertinent to the question. Respond concisely and in a step-by-step manner.~~ Query: [query]. Please provide detailed findings in response to the query: \\ 
\hline\hline
\textbf{Query: find the key reference for the following paper }\\ 
\arrayrulecolor{black}\hline
\textbf{Response from Our system }\\
Key Reference: [Training a helpful and harmless assistant with reinforcement learning from human feedback, Yuntao Bai; Andy Jones; Kamal Ndousse; Amanda Askell; Anna Chen; Nova Dassarma; Dawn Drain; Stanislav Fort; Deep Ganguli; Tom Henighan, 2022]

Why this reference is important to the document:
This paper is crucial to the document because it presents a significant prior work on the training and alignment of large language models, which is directly applicable to the authors' study. The referenced paper investigates reinforcement learning from human feedback as a method to align language models with human values, which is a strategy the authors are further exploring in their work with LIMA. It provides a foundational understanding of how language models can be trained to be both helpful and non-harmful, which is crucial for the development and evaluation of LIMA.

Main Research Direction:
The core research direction of the key reference is the enhancement of the efficacy and safety of language models via training techniques. The authors employ reinforcement learning from human feedback to align the model's outputs more closely with human expectations and preferences. This approach involves training a model to predict the actions that a human demonstrator would take, and then refining it based on a reward model derived from comparisons of different potential model outputs. This method is intended to make the model more useful and less likely to generate harmful or inappropriate content. The research provides valuable insights into how to train and fine-tune language models to make them more effective and safe for practical applications.
token cost : 1397
\\ 
\hline
\textbf{Response from chatpdf.com} \\
The key reference for the paper mentioned in the given texts is [Kirstain et al., 2021]. \\ 
\hline
\textbf{Response from Claude+ 100K (GT of Evaluation.)~} \\
Based on an analysis of the references provided, References 5 and 6 appear to be the key references for this paper:

References 5: Scaling language modeling with pathways
Aakanksha Chowdhery, Sharan Narang, Jacob Devlin, et al.

References 6: Scaling instruction-finetuned language models

Mostafa Dehghani, Siddhartha Brahma, et al.

These two references describe scaling up and fine-tuning large language models, which is directly relevant to the focus of the paper on fine-tuning a 65B parameter language model. Several other references discuss specific language models (GPT-4, Bard, DaVinci003, etc.) that the authors compare their model to, but References 5 and 6 seem to provide the most direct background on the technique of fine-tuning large language models, which is the core method used in the paper.

The other references provide relevant background on specific techniques (like reinforcement learning from human feedback), specific datasets, or are used to justify certain claims. But References 5 and 6 appear to be the most fundamental in informing the authors' approach of fine-tuning a large language model to demonstrate their hypotheses about alignment.

Does this match your assessment of the key references? Let me know if you have any questions or require any clarification. \\ 
\hline
\textbf{Response from MPT-30B-Chat} \\
Title: Transformer-based Language Models: A Survey of the State-of-the-Art

Author: Zhiyang Peng, Xipeng Qiu, photosynthetic, Xuebo Liu, Qingquan Song, Jun Zhou, year: 2022

This reference is a survey paper that provides a comprehensive overview of transformer-based language models, which is relevant to Section 2.1 as it covers the architecture and components of

transformer-based models such as BERT, GPT, and RoBERTa. The paper discusses the training objectives, pre-training tasks, and fine-tuning methods of these models, as well as their applications in

various natural language processing tasks. The reference also discusses the limitations and future directions of transformer-based language models.
 \\ 
\hline
\textbf{Response from Vicuna-33B} \\
1 Gehrmann, J., \& Ayyash, A. (2022). The benefits of pressure.

2 D'mello, S. K., \& Graesser, A. C. (2012). \\ 
\hline
\textbf{Response from Vicuna-13B} \\
The main research direction of the key references in this document can
be categorized into several areas:

1 Improving the performance of language models: Many of the references focus on improving the performance
of language models by fine-tuning them on specific tasks or by using unsupervised or semi-supervised

learning methods. For example, reference [1] proposes a method for fine-tuning language models on

specific tasks, while reference [8] describes a method for improving language models with almost no

human labor.

2 Enabling language models to follow instructions: Several references focus on enabling language models

to follow instructions given in natural language. For example, reference [17] describes a method for

training language models to follow instructions with human feedback, while reference [20] describes an
instruction-following language model called Stanford Alpaca.

3 Controllable language generation: Several references describe methods for generating text that can be

controlled by the user, either by providing input or by specifying the desired output. For example,

reference [10] describes a language model called Ctrl that can generate text based on a set of

instructions, while reference [24] describes a method for finetuning language models to make them

zero-shot learners.

4 Democratizing large language model alignment: Some references focus on making it easier for researchers
and practitioners to use large language models for a variety of tasks. For example, reference [13]

describes a method for democratizing large language model alignment, while reference [25] describes a

method for eliciting reasoning in large language models through chain-of-thought prompting.

Overall, the key references in this document reflect the ongoing efforts of researchers and practitioners
to improve the performance and capabilities of language models, and to make them more accessible and

useful for a wide range of applications. \\ 
\hline
\textbf{Response from Vicuna-7B} \\
I'm sorry, but I cannot provide an explanation without knowing the

specific key reference you are referring to. Please provide the key reference in the format specified so

that I can assist you better.\\
\bottomrule
\end{longtable*}


\end{document}